%% file: example_paper.tex
\documentclass{article}

\usepackage{microtype}
\usepackage{graphicx}
\usepackage{subfigure}
\usepackage{booktabs} 

\usepackage{hyperref}

\usepackage[accepted]{icml2024}

\usepackage{amsmath}
\usepackage{amssymb}
\usepackage{mathtools}
\usepackage{amsthm}
\usepackage{xspace}

\usepackage[capitalize,noabbrev]{cleveref}

\theoremstyle{plain}
\newtheorem{theorem}{Theorem}[section]
\newtheorem{proposition}[theorem]{Proposition}
\newtheorem{lemma}[theorem]{Lemma}
\newtheorem{corollary}[theorem]{Corollary}
\theoremstyle{definition}
\newtheorem{definition}[theorem]{Definition}
\newtheorem{assumption}[theorem]{Assumption}

\usepackage[disable,textsize=tiny]{todonotes}
\def\shownotes{0}  
\ifnum\shownotes=1
\newcommand{\authnote}[2]{[#1: #2]}
\else
\newcommand{\authnote}[2]{}
\fi

\input{macros}

\input{std_macros}

\icmltitlerunning{Connect Later: Improving Fine-tuning for Robustness with Targeted Augmentations}

\begin{document}

\twocolumn[
\icmltitle{Connect Later: Improving Fine-tuning for Robustness with Targeted Augmentations}




\begin{icmlauthorlist}
\icmlauthor{Helen Qu}{penn}
\icmlauthor{Sang Michael Xie}{stanford}
\end{icmlauthorlist}

\icmlaffiliation{penn}{University of Pennsylvania}
\icmlaffiliation{stanford}{Stanford University}

\icmlcorrespondingauthor{Helen Qu}{helenqu@sas.upenn.edu}

\icmlkeywords{robustness, machine learning, pretraining, fine-tuning}

\vskip 0.3in
]



\printAffiliationsAndNotice{\icmlEqualContribution} 

\begin{abstract}
Models trained on a labeled source domain often generalize poorly when deployed on an out-of-distribution (OOD) target domain. In the domain adaptation setting where unlabeled target data is available, self-supervised pretraining (e.g., contrastive learning or masked autoencoding) is a promising method to mitigate this performance drop. Pretraining depends on generic data augmentations (e.g., cropping or masking) to learn representations that generalize across domains, which may not work for all distribution shifts. In this paper, we show on real-world tasks that standard fine-tuning after pretraining does not consistently improve OOD error over simply training from scratch on labeled source data. To better leverage pretraining for distribution shifts, we propose the Connect Later framework, which fine-tunes the model with \emph{targeted augmentations} designed with knowledge of the shift. Intuitively, pretraining learns good representations within the source and target domains, while fine-tuning with targeted augmentations improves generalization across domains. Connect Later achieves state-of-the-art OOD accuracy while maintaining comparable or better in-distribution accuracy on 4 real-world tasks in wildlife identification (\iwildcam), tumor detection (\camelyon), and astronomy (\classification, \redshifts).
\end{abstract}

\section{Introduction}

In the real world, machine learning models are often deployed on data that differ significantly from training data \citep{quinonero2009dataset, koh2021wilds}. We focus on unsupervised domain adaptation \citep{shimodaira2000improving,blitzer2006domain,sugiyama2007covariate}, where we have labeled data from a source domain and unlabeled data from a target domain. We aim to learn a model that generalizes well to these out-of-distribution (OOD) target domain inputs.
A real-world example is wildlife identification, where the task is to identify animal species from static camera trap images.
However, human labels are only available for images from a small subset of these cameras, which may not be representative of the habitats and characteristics of unlabeled camera images.

Pretraining on broad unlabeled data has shown promising results on improving OOD error in real-world problems~\citep{caron2020swav, shen2022connect, radford2021clip, sagawa2022uwilds}.
In particular, contrastive pretraining has been shown to learn representations that transfer well across domains~\citep{shen2022connect,haochen2022beyond}.
In contrast to conventional domain adaptation methods that focus on learning domain-invariant features~\citep{ganin2016domain,kang2019contrastive,tzeng2017domain,saenko2010adapting,sun2016return,hoffman2018cycada}, contrastive pretraining learns representations that are not domain-invariant, but instead decompose the class and domain information, facilitating transfer across domains~\citep{shen2022connect}. 
A favorable decomposition depends on the generic data augmentations used during contrastive pretraining to align representations across domains.
Intuitively, augmented (e.g. masked or cropped) source and target inputs should be more likely to look similar if they are from the same class (e.g., cropping out the face of a lion in different habitats) than from different classes (e.g., no body parts of elephants and lions are alike).
However, these generic augmentations may not be suitable for all distribution shifts.

In this paper, we find on real-world benchmarks that standard fine-tuning after contrastive pretraining is not always effective for improving OOD error over purely supervised learning from scratch with labeled source data (Section~\ref{sec:pretraining}).
On the other hand, supervised learning with \emph{targeted augmentations} \citep{gao2023targeted} designed for the distribution shift improves OOD error over the supervised learning baseline on all datasets without access to any target unlabeled data.
Thus, pretraining does not always learn representations that transfer across domains with standard fine-tuning.

\begin{figure}
    \centering
    \includegraphics[scale=0.18]{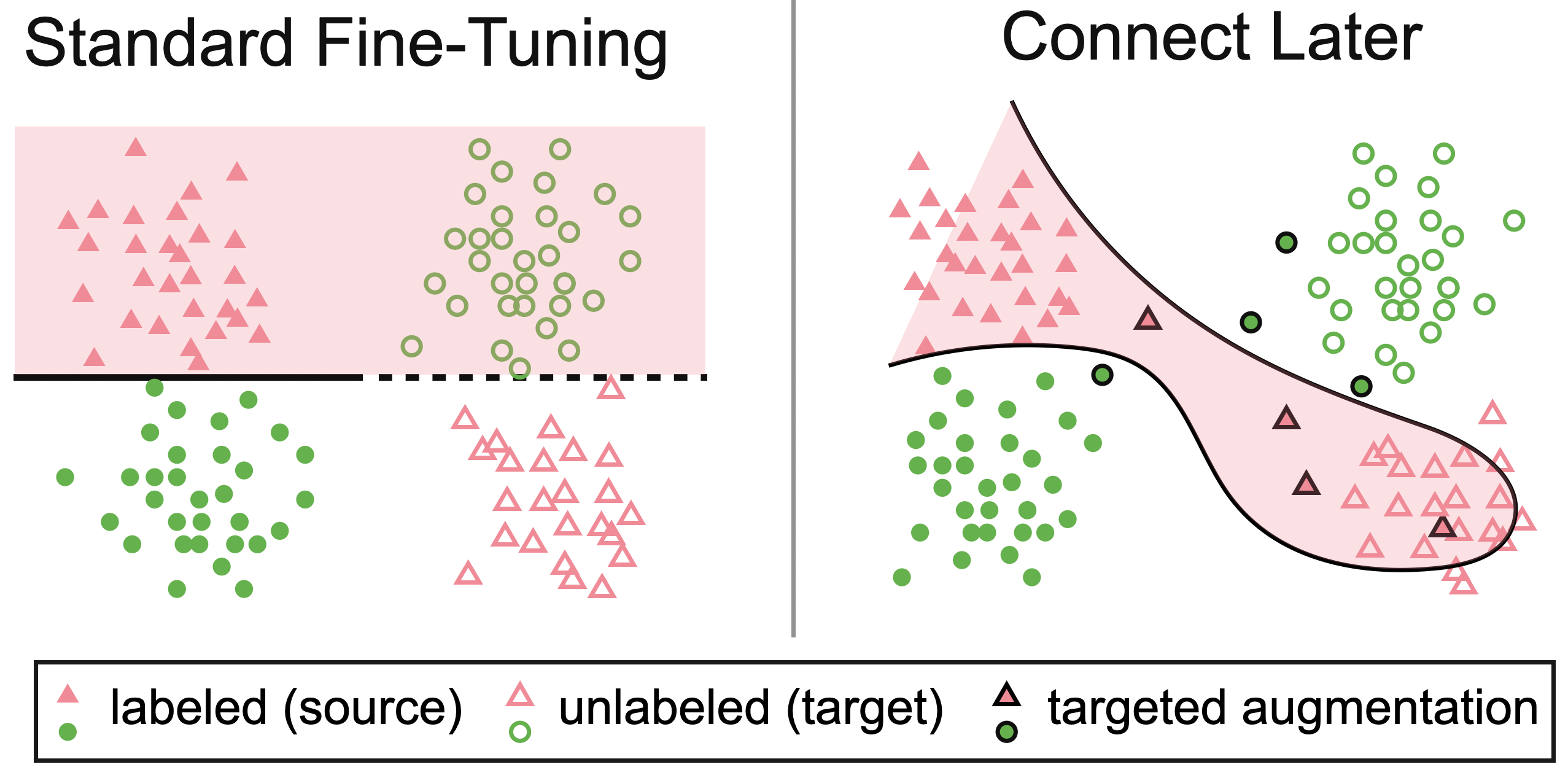}
    \caption{Overview of the Connect Later framework applied to a toy binary classification problem with two domains (filled and unfilled points), showing the representations from contrastive pretraining in $\R^2$. \textbf{(Left)} After contrastive pretraining with generic augmentations, the classes within each domain are linearly separable in representation space. Since the domains are far apart in input space, generic augmentations may misalign the pretrained representations. In this case, a classifier (with a linearly extrapolating decision boundary, dashed and solid line) learned on labeled source data will misclassify the target data. \textbf{(Right)} Connect Later employs targeted augmentations (filled points with black border) designed with knowledge of the distribution shift to connect the domains better, resulting in a classifier that generalizes well to the target domain.}
    \label{fig:overview}
\end{figure}

To better leverage pretraining for domain adaptation, we propose the Connect Later framework (Figure~\ref{fig:overview}): after pretraining with generic augmentations, fine-tune with targeted augmentations (Section~\ref{sec:connect-later}).
Intuitively, pretraining learns good representations within each domain, while targeted augmentations incorporate domain knowledge to improve generalization across domains.
Through both empirical and theoretical examples, we show that Connect Later generalizes well to the target domain even in scenarios where pretraining alone produces minimal OOD performance improvements.
We provide a general methodology for constructing these targeted augmentations by matching augmented inputs to the target distribution on a feature space where the domains differ.

We evaluate our framework on 4 real-world datasets: wildlife identification \citep[\iwildcam,][]{beery2020iwildcam, sagawa2022uwilds}, tumor detection \citep[\camelyon,][]{bandi2018detection, sagawa2022uwilds} and 2 astronomical time series tasks, \classification and \redshifts, which we curate from \citet{theplasticcteam2018photometric}.
In Section~\ref{sec:results}, we show that Connect Later improves OOD performance over \sft or supervised learning with targeted augmentations across all datasets.
Although our understanding stems from contrastive learning, we empirically apply Connect Later to better leverage pretrained representations from both masked autoencoding and contrastive learning.
Connect Later achieves the state-of-the-art on three benchmarks, improving OOD accuracy on \classification by 3\% \citep{Boone_2019}, \iwildcam with ResNet-50 by 0.9\%, and \camelyon with DenseNet121 by 1.1\%. We also contribute the \redshifts dataset, on which Connect Later outperforms the best baseline by 11\% relative improvement. 

\section{Setup}
We consider a prediction problem from an input space $\sX$ to a label space $\sY$, where $\sY = \{1,\dots,k\}$ for classification and $\sY \in \R$ for regression.

\paragraph{Domain adaptation.}
Let $P_S$ and $P_T$ be the source and target input distributions over $\sX$, respectively.
We consider unsupervised domain adaptation, where we have access to source inputs $x \sim P_S$, with corresponding labels $y\in\sY$ sampled from the label distribution $\pstar(\cdot \mid x)$, along with unlabeled target inputs sampled from the target distribution $P_T$. 
Let the unlabeled distribution $P_U = \beta P_S + (1-\beta) P_T$ be a mixture of the source and target, where $\beta \in [0,1]$. In practice, $P_U$ may also be a broader unlabeled distribution.
The goal is to learn a model $f: \sX \rightarrow \sY$ that minimizes error on the target domain $L_T(f) = \E_{x\sim P_T, y\sim \pstar(\cdot \mid x)}[\loss(f(x), y)]$. For example, $\loss:\sY\times \sY \rightarrow \R$ is the 0-1 loss in classification and squared loss in regression.

\paragraph{Augmentations.}
Augmented inputs $\inputxp \in \sX$ are drawn from an augmentation distribution $\sA(\cdot | x)$, given an input $x \in \sX$. 
Training with augmented inputs is often used to improve robustness \citep{hendrycks2019augmix, hendrycks2020many} and is crucial to contrastive pretraining \citep{caron2020swav,shen2022connect, devlin2019bert}.
In this work, we define two distinct augmentation distributions, $\aug$ and $\augft$, for the pretraining and fine-tuning steps, respectively. 
Typically, the pretraining augmentations $\aug$ are generic transformations, such as random cropping in vision or masking in NLP \citep{caron2020swav,chen2020simclr,he2020moco,radford2021clip,shen2022connect,he2022mae,devlin2019bert}. 
Fine-tuning augmentations $\augft$ have not been studied extensively and are typically also generic or simply the identity transformation \citep{sagawa2022uwilds, devlin2019bert}.

\paragraph{Contrastive pretraining for domain adaptation.}
Contrastive pretraining for domain adaptation consists of two steps: self-supervised pretraining on unlabeled data, then supervised fine-tuning on labeled source data \citep{shen2022connect}.
For simplicity below, we consider the population objectives.
Contrastive learning aims to learn an encoder which maps augmented views of the same input to similar features (``positive pairs'') and views of different inputs to dissimilar features (``negative pairs''), according to some distance metric. Formally, let $\pospairdist(\inputx, \posx)=\E_{\bar{\inputx}\sim \unlabeldist}[\aug(\inputx \mid \bar{\inputx})\aug(\posx \mid \bar{\inputx})]$ be the distribution over positive pairs, which are augmentations of a single input $\bar{\inputx}$. 
We pretrain an encoder $\encoder:\sX \rightarrow \R^\embeddim$ to minimize the distance $\dattract$ between positive pair embeddings and maximize the distance $\drepel$ between negative pair embeddings:
\begin{equation}
\begin{split}
\label{eqn:pretrain_objective}
\sL_{\text{pretrain}}(\encoder) = \E_{(\inputx,\posx)\sim \pospairdist}[\dattract(\encoder(\inputx), \encoder(\posx))] - \\
\E_{\inputx,\inputxp\sim \unlabeldist}[\drepel(\encoder(\inputx), \encoder(\inputxp))].
\end{split}
\end{equation}
The output of the pretraining step is a pretrained encoder $\empencoder=\argmin_\encoder \sL_{\text{pretrain}}(\encoder)$.


Fine-tuning then learns a prediction head $\head:\R^\embeddim\rightarrow \R^n$ 
(for regression, we let $n=1$) on top of the pretrained encoder using labeled source data with the objective
\begin{align}
\label{eqn:ft_objective}
    \sL_{\text{ft}}(\head) = \E_{\inputx \sim P_S, y\sim \pstar(\cdot \mid \inputx), \inputxp \sim \augft(\cdot | \inputx)} [\lossft(\head(\empencoder(\inputxp)),\;y;\; \theta)]
\end{align}
where $\lossft: \R^n \times \sY \rightarrow \R$ is a fine-tuning objective such as softmax cross entropy loss for classification or squared error for regression. The learned head is $\emphead=\argmin_{\head} \sL_{\text{ft}}(\head)$. In practice, we jointly fine-tune the head $\head$ and the encoder $\empencoder$.

\paragraph{\Sft.}
We refer to \textbf{\sft} as the pretraining+fine-tuning procedure where $\augft(x'\mid x)=1$ if $x'=x$ (no fine-tuning augmentations).
In our experiments, the standard fine-tuning procedure is linear probing then fine-tuning (LP-FT)~\citep{kumar2022finetuning}, which has been shown to improve ID and OOD performance over vanilla fine-tuning. In LP-FT, we first learn a linear predictor on top of frozen pretrained features before fine-tuning all the parameters jointly.

\paragraph{ERM with augmentations.}
As a baseline, we consider empirical risk minimization (ERM) with data augmentation, which optimizes the fine-tuning objective (Equation~\ref{eqn:ft_objective}) on labeled source data with randomly initialized parameters.
In this paper, we refer to \textbf{ERM} as the instantiation where $\augft(x'\mid x) = 1$ if $x'=x$ (no augmentations) and \textbf{ERM + targeted augmentations} as the instantiation with $\augft$ that is designed with knowledge of the distribution shift.


\section{Pretraining may not improve OOD performance}
\label{sec:pretraining} 

\begin{table}[]
    \centering
    \caption{Contrastive pretraining with standard fine-tuning substantially improves OOD performance for \camelyon but is not very effective for \iwildcam. Results are averaged over 15 trials for \iwildcam and 20 trials for \camelyon, and we report the 95\% confidence intervals on each mean estimate.}
    \resizebox{0.47\textwidth}{!}{
        \begin{tabular}{lcccc}
            \toprule
            & \multicolumn{2}{c}{iWildCam (Macro F1, $\uparrow$)} & \multicolumn{2}{c}{Camelyon17 (Avg Acc, $\uparrow$)} \\
            & ID Test  & OOD Test & ID Val & OOD Test \\
            \midrule
            ERM & $46.4 \pm 0.5$ & $30.4 \pm 0.6$ & $89.3 \pm 0.9$ & $65.2 \pm 1.1$\\
            {Standard fine-tuning} & $46.4 \pm 0.8$ & $31.2 \pm 0.6$ & $92.3 \pm 0.2$ & $91.4 \pm 0.9$ \\
            \bottomrule
        \end{tabular}
    }
    \label{tbl:observations}
\end{table}

We compare ERM and \sft on two benchmark datasets, \iwildcam (wildlife species identification) and \camelyon (tumor detection).
In Table~\ref{tbl:observations}, we show that \sft on a model pretrained using SwAV contrastive learning~\citep{caron2020swav} makes minimal gains over ERM on \iwildcam ($46.4 \rightarrow 46.4$ ID, $30.4 \rightarrow 31.2$ OOD) compared to \camelyon ($89.3 \rightarrow 92.3$ ID, $65.2 \rightarrow 91.4$ OOD).
This result runs contrary to prior work demonstrating that contrastive pretraining is an effective domain adaptation method \citep{caron2020swav, shen2022connect, radford2021clip, sagawa2022uwilds}.
We hypothesize that the generic pretraining augmentations connect the domains better for some tasks and distribution shifts than others.

\paragraph{Simple example with misaligned connectivity structure.}
To understand this phenomenon, we provide a simple binary classification example of when contrastive pretraining fails for domain adaptation, following a similar augmentation graph construction to~\citet{shen2022connect}, in Appendix~\ref{app:simple_example}.
When the connectivity structure misaligns the source and target domains, such that examples from the same class are less ``connected'' than examples from different classes across the domains, a linear classifier trained on these pretrained representations will not transfer from source to target.
This could happen, for example, when the source and target are far apart in input space and connectivity is low between examples from the same class across different domains.

\subsection{Robustness gains from pretraining depend on dataset connectivity}
To better understand why contrastive pretraining performs differently on these two datasets, we empirically evaluate the connectivity measures for \iwildcam and \camelyon. 
We follow \citet{shen2022connect} and work in the augmentation graph setting, where nodes are inputs and edge weights are the positive-pair probabilities given by $\pospairdist$. 
We define connectivity between a class-domain pair $((y_1, d_1), (y_2, d_2))$ under four scenarios:
\begin{align}
    \begin{cases}
        \pairprobr & y_1 = y_2, d_1 = d_2~\text{~~(same class, same domain)}\\
        \pairproba & y_1 = y_2, d_1 \neq d_2\text{~~(same class, different domain)}\\
        \pairprobb & y_1 \neq y_2,d_1 = d_2\text{~~(different class, same domain)}\\
        \pairprobg & y_1 \neq y_2, d_1 \neq d_2\text{~~(different class and domain)}\\  
    \end{cases},
\end{align}
where each value is an average edge weight over the edges that satisfy each case.
\citet{shen2022connect} show in simple augmentation graphs that contrastive pretraining theoretically learns transferable representations when $\pairproba > \pairprobg$ and $\pairprobb > \pairprobg$, and that the ratios $\frac{\pairproba}{\pairprobg}$ and $\frac{\pairprobb}{\pairprobg}$ empirically correlate well with OOD accuracy.
Intuitively, the pretraining augmentations are less likely to change both the domain and class of an input than changing just domain or just class.

\begin{table}[]
    \centering
    \caption{Empirically estimated connectivity measures for \iwildcam and \camelyon. From~\citet{shen2022connect}, contrastive pretraining theoretically learns transferable representations for UDA when both across-domain ($\alpha$) and across-class ($\beta$) connectivity is greater than across-both ($\gamma$). In \iwildcam, $\gamma > \beta$, violating the condition, while \camelyon satisfies the condition.}
    \resizebox{0.4\textwidth}{!}{
        \begin{tabular}{lccc}
            \toprule
             & $\alpha$ & $\beta$ & $\gamma$ \\ \midrule
            \iwildcam & 0.116 & 0.071 & 0.076 \\
            \camelyon & 0.16 & 0.198 & 0.152 \\ 
            \bottomrule
        \end{tabular}
    }
    \label{tbl:connectivity}
\end{table}

\paragraph{Empirical evaluations of connectivity.}
We empirically evaluate the connectivity measures for \iwildcam and \camelyon following \citet{shen2022connect}. Using augmented inputs from 2 class-domain pairs, we train a binary classifier to predict the class-domain pair of each input, and interpret the test error of the classifier as an estimate for connectivity.
We average each connectivity value over 15 class-domain pairs (see Appendix~\ref{app:connectivity} for details).
Our results, summarized in Table~\ref{tbl:connectivity}, show that \iwildcam connectivity measures violate the condition for contrastive pretraining in the UDA setting, since across-both connectivity $>$ across-class ($\gamma > \beta$). 
This finding is consistent with our observation that contrastive pretraining is far less effective for \iwildcam compared to \camelyon, and further underscores the need for domain adaptation methods that correct the misaligned connectivity structure.

\section{Connect Later: Pretrain First, Targeted Augmentations Later}
\label{sec:connect-later}

Even when generic augmentations applied during pretraining misalign the connectivity structure, the pretrained representations are still useful since the classes are linearly separable \textit{within} each domain.
How do we leverage these pretrained representations when they may not transfer well across domains?
In this work, we propose the Connect Later framework (Figure~\ref{fig:overview}):
\begin{enumerate}
  \item Pretrain on unlabeled data with generic augmentations as in Equation~\ref{eqn:pretrain_objective}, producing a pretrained encoder $\empencoder$. 
  \item Design a targeted augmentation $\augft$ (discussed below) and use augmented source data to fine-tune the pretrained encoder $\empencoder$ jointly with a prediction head $h$ as in Equation~\ref{eqn:ft_objective}.
\end{enumerate}
While our intuition about pretraining for domain adaptation stems from \citet{shen2022connect}, we show that applying targeted augmentations at fine-tuning time is sufficient for good generalization to the target domain even when the pretrained representations transfer across domains poorly. 
This allows us to reuse pretrained models for multiple downstream tasks.

\paragraph{Simple example where Connect Later achieves 0 OOD error.} In our simple binary classification example in Appendix~\ref{app:simple_example}, we show that when the connectivity structure is misaligned, both \sft with contrastive pretraining and ERM + targeted augmentations have high OOD error, while Connect Later achieves 0 OOD error.
In this setting, ERM with targeted augmentations is unable to achieve 0 OOD error since some target inputs are ``unreachable'' via targeted augmentations of source inputs. The pretraining step in Connect Later uses unlabeled target data to learn representations where label information from source data can propagate to all target inputs.

\subsection{Real-world examples of targeted augmentations} 
\begin{figure}[t]
    \centering
    \includegraphics[scale=0.22]{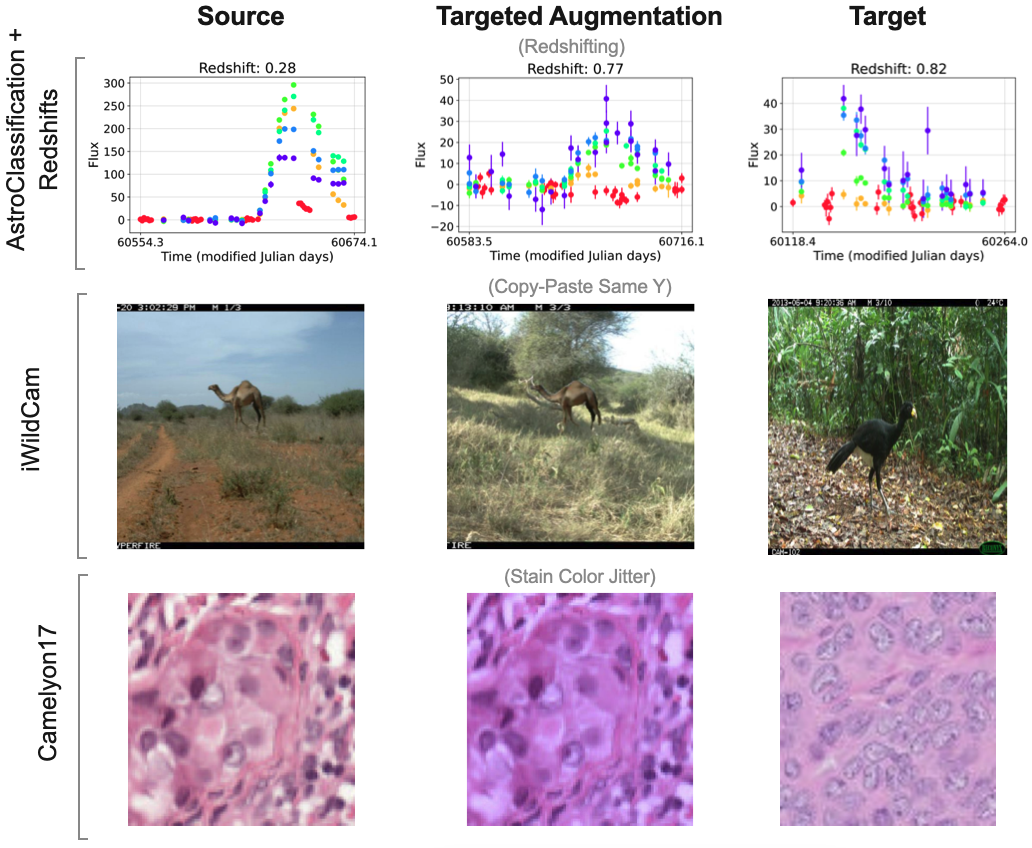}
    \caption{Examples from the source dataset (left), an augmented version of the source example (middle), and the target dataset (right) for each of our tasks. (\textbf{Top row}) The \classification and \redshifts tasks focus on time-varying astronomical objects observed in multiple wavelength ranges, plotted here as a multicolored time-series with each color corresponding to the wavelength range of the measurement. The redshifting augmentation simulates placing source objects at a higher redshift to better match the redshift distribution of the target dataset. The flux errors and flux values of the augmented example (middle) show much better resemblance to the target example. (\textbf{Middle row}) We randomize the habitat background by applying the Copy-Paste Same Y augmentation for \iwildcam (\iwildcam image examples shown here are from \citet{gao2023targeted}). (\textbf{Bottom row}) Stain Color Jitter alters the overall color of source images in \camelyon to improve performance on images from unseen hospitals.}
    \label{fig:datasets}
\end{figure}

We design targeted augmentations for 4 real-world tasks: wildlife identification, tumor detection, and astronomical time-series classification and redshift prediction. We show examples from the source, augmented, and target datasets for these tasks in Figure~\ref{fig:datasets}. 

\paragraph{Wildlife species classification (\iwildcam).} For \iwildcam \citep{beery2020iwildcam,sagawa2022uwilds}, the task is to identify the wildlife species from static camera trap images. 
These cameras are placed in a wide variety of environments, which all have unique habitat conditions (e.g., African savannah vs. tropical rainforest) and camera characteristics (e.g., angles, resolutions).
\begin{itemize}
    \item \textbf{Source:} 243 camera traps
    \item \textbf{Target:} 48 unseen camera traps
    \item \textbf{Targeted Augmentation:}
We augment the labeled dataset with the Copy-Paste Same Y algorithm, which uses image segmentation to copy-paste the animal onto different background images from cameras that have observed the same species \citep{gao2023targeted}. 
    \item \textbf{Task:} 182-class wildlife species classification
\end{itemize}

\paragraph{Tumor detection (\camelyon).} The task in \camelyon \citep{bandi2018detection} is to classify whether a patch of a histopathology slide contains a tumor. These slides are contributed from multiple hospitals, which use different stain colors and also vary in distributions of patient cancer stage.
\begin{itemize}
    \item \textbf{Source:} Hospitals 1-3.
    \item \textbf{Target:} Hospitals 4 and 5.
    \item \textbf{Targeted Augmentation:}
We augment the labeled dataset with the Stain Color Jitter algorithm, which jitters the color of the slide image in the hematoxylin and eosin staining color space \citep{tellez2018whole}. 
    \item \textbf{Task:} Determine if a slide contains a tumor.
\end{itemize}

\paragraph{Astronomical object classification (\classification).}
Astronomical object classification \citep{Boone_2019, allam2022paying} involves predicting the object type (e.g., type II supernova) from a time series of an object's brightness at multiple wavelengths. 
Expert labeling is only available for nearby objects, which are brighter and have different properties than distant objects (see Appendix~\ref{app:data} for details).

\begin{itemize}
    \item \textbf{Source:} Time-series of bright, nearby objects with expert labels
    \item \textbf{Target:} Time-series of all observed objects from the telescope, often faint and distant (higher redshift). Follow-up observation, which is required for expert labeling, is too expensive for these objects.
    \item \textbf{Targeted Augmentation:} 
We augment the labeled dataset by redshifting each object, i.e., simulating its observed properties as if it were further away (details in Appendix~\ref{app:targeted-augs}).
    \item \textbf{Task:} 14-class astronomical object classification
\end{itemize}

\paragraph{Redshift regression (\redshifts).}
Similar to object type, redshift information is also available only for bright, nearby objects. We predict the scalar redshift value of each object and minimize mean squared error. \redshifts is a new dataset that we contribute as part of this work.

\begin{itemize}
    \item \textbf{Source:} Time-series of bright, nearby labeled objects.
    \item \textbf{Target:} Time-series of all observed objects from the telescope, often faint and distant (higher redshift).
    \item \textbf{Targeted Augmentation:} Redshifting (same as \classification, Appendix~\ref{app:targeted-augs}).
    \item \textbf{Task:} Redshift regression
\end{itemize}

\subsection{Designing targeted augmentations}
Targeted augmentations offer the opportunity to incorporate domain knowledge to improve generalization performance. We provide a general methodology for designing targeted augmentations based on matching the target distribution on a feature space:

\begin{enumerate}
    \item Identify a feature space $\sZ$. We assume that we can label $z \in \sZ$ for each input and that the source and target domains largely differ on this feature space. One such example is the space of spurious, domain-dependent features (e.g., camera angle or resolution for \iwildcam), which is the approach followed by~\citet{gao2023targeted}. 
    \item Fit a transformed feature distribution $\hat{p_T}(\znew | z)$ to the target feature distribution. 
    \item Create a transformation distribution $T(x' | x, \znew)$ where $x'$ is the augmented version of $x$ with $z=\znew$. In this paper, we define $T$ with domain knowledge.
    \item Given an input $x$, generate augmentations by sampling a new feature $\znew$ from $\hat{p_T}(\znew \mid z)$, then sampling an augmentation from $T(x' | x, \znew)$. The resulting targeted augmentation probabilities are $\augft(x' \mid x) = \sum_{\znew} T(x' \mid x, \znew) \hat{p_T}(\znew \mid z)$.
\end{enumerate}

\paragraph{Targeted augmentation example.}
We follow the procedure outlined above to design a targeted augmentation for \classification and \redshifts (see Appendix~\ref{app:targeted-augs} for further details).
\begin{enumerate}
    \item The source and target domains have different redshift distributions, so we identify this scalar feature as $z$.
    \item We roughly fit the target redshift distribution within a reasonable range of the original redshift $z$, such that $\hat{p_T}(\znew \mid z)$ is distributed as $\text{loguniform}(0.95z,\; \text{min}(1.5(1+z)-1, \;5z))$, following \citet{Boone_2019}.
    \item We define a transformation distribution $T(x' | x, \znew)$, where $x$ is a time-series of flux values at multiple wavelengths and $\znew$ is a new redshift value to transform to. We first fit a Gaussian process that models $x$ as a function of time and wavelength. Given $\znew$, we rescale the timestamps and wavelengths of the original input to account for the physical effects of the new redshift value. Then, we sample $\tilde{x'}$ from the Gaussian process at these new timestamps and wavelengths. Finally, we produce the transformed input $x'$ by scaling the flux values to account for $\znew$.
    \item We sample $\znew$ from $\hat{p_T}(\znew \mid z)$ and then sample augmentations $x'$ from $T(x' | x, \znew)$.
\end{enumerate}

\section{Experiments}
\label{sec:results}
We empirically test Connect Later with contrastive pretraining (\iwildcam, \camelyon) as well as pretraining with masked autoencoding (\classification, \redshifts) to demonstrate Connect Later as a general fine-tuning method.
We note that masked autoencoding has been linked to contrastive learning in \citet{zhang2022mask}, which shows that the masked autoencoding objective upper bounds the contrastive loss between positive pairs -- thus, masked autoencoding implicitly aligns the positive pairs induced by the masking augmentations.

\paragraph{Training procedure.} 

For \iwildcam, we use a ResNet-50 model pretrained on unlabeled ImageNet data with SwAV contrastive learning~\citep{caron2020swav}.
We use a DenseNet121 pretrained on unlabeled data from \citet{sagawa2022uwilds} with SwAV for \camelyon.
We pretrain with masked autoencoding for \classification and \redshifts by masking 60\% of observations from each light curve (Appendix~\ref{app:experiments}). 
The same pretrained model is used for both tasks to demonstrate the reusability of pretrained features.
We fine-tune the pretrained models with linear probing then fine-tuning~\citep[LP-FT,][]{kumar2022finetuning}, which has been shown to improve OOD performance.

\paragraph{Baselines.}
We evaluate our framework against three baselines: ERM, ERM+targeted augs, and \sft. We include Avocado \citep{Boone_2019}, the state-of-the-art model for \classification. We also include a self-training baseline for \classification and \redshifts, which has been shown to perform well on some real-world datasets \citep{sagawa2022uwilds}. For the self-training baseline, we pseudo-label the target dataset with a trained ERM+targeted augs model, then combine with the labeled source dataset and apply the targeted augmentation for training.
We include additional domain adaptation baselines for \iwildcam and \camelyon: domain-adversarial neural networks~\citep[DANN,][]{ganin2016domain}, correlation alignment~\citep[CORAL,][]{sun2016return}, Noisy Student \citep{xie2020selftraining}, and ICON\footnote{\label{icon}https://github.com/a-tea-guy/ICON}.

\subsection{Main results}
\label{sec:main-results}

\begin{table*}[]
    \centering
    \caption{ID and OOD accuracy (\%) for \classification and RMSE for \redshifts of each method. Results are averaged over 5 trials and rows with means within 1 STD of the best mean are bolded.}
    \resizebox{0.7\textwidth}{!}{
        \begin{tabular}{lcccc}
            \toprule
            & \multicolumn{2}{c}{AstroClassification} & \multicolumn{2}{c}{Redshift}\\
            & ID Test Acc ($\uparrow$) & OOD Acc ($\uparrow$) & ID Test RMSE ($\downarrow$) & OOD RMSE ($\downarrow$) \\
            \midrule
            
            ERM & $71.59 \pm 1.10$ & $61.26 \pm 1.10$ & $0.274 \pm 0.016$ & $0.320 \pm 0.009$ \\
            \Sft & $78.84 \pm 0.97$ & $67.84 \pm 0.70$ & $\mathbf{0.246 \pm 0.015}$ & $0.277 \pm 0.004$ \\
            ERM + targeted augs & $68.75 \pm 0.95$ & $67.54 \pm 0.32$ & $0.310 \pm 0.006$ & $0.286 \pm 0.007$ \\
            Self-Training & $77.72 \pm 0.59$ & $65.15 \pm 0.67$ & $0.304 \pm 0.010$ & $0.289 \pm 0.003$ \\
            Avocado \citep{Boone_2019} & - & $77.40$ & - & - \\
            Connect Later & $\mathbf{80.54 \pm 1.20}$ & $\mathbf{79.90 \pm 0.60}$ & $\mathbf{0.256 \pm 0.005}$ & $\mathbf{0.247 \pm 0.005}$ \\
            \bottomrule
        \end{tabular}
    }

    \label{tbl:main-astro}
\end{table*}

\begin{table*}[]
    \centering
    \caption{ID and OOD performance for each method on \iwildcam and \camelyon. Results are averaged over 15 trials for \iwildcam and 20 trials for \camelyon, and we report 95\% confidence intervals on each mean estimate. Rows with means within 1 interval of the best mean are bolded.}
    \resizebox{0.65\textwidth}{!}{
        \begin{tabular}{lcccc}
            \toprule
            & \multicolumn{2}{c}{iWildCam (Macro F1, $\uparrow$)} & \multicolumn{2}{c}{Camelyon17 (Avg Acc, $\uparrow$)} \\
            & ID Test  & OOD Test & ID Val & OOD Test \\
            \midrule
            ERM & $46.4 \pm 0.5$ & $30.4 \pm 0.6$ & $89.3 \pm 0.9$ & $65.2 \pm 1.1$ \\
            \Sft & $46.4 \pm 0.8$ & $31.2 \pm 0.6$ & $92.3 \pm 0.2$ & $91.4 \pm 0.9$\\
            ERM + targeted augs & $\mathbf{51.4 \pm 0.6}$ & $36.1 \pm 0.7$ & $96.7 \pm 0.0$ & $90.5 \pm 0.4$\\
             DANN ~\citep{sagawa2022uwilds} & $48.5 \pm 3.2$ & $31.9 \pm 1.6$ & $86.1 \pm 1.3$ & $64.5 \pm 1.2$ \\
            CORAL ~\citep{sagawa2022uwilds} & $40.5 \pm 1.6$ & $27.9 \pm 0.5$ & $92.3 \pm 0.7$ & $62.3 \pm 1.9$\\
            Noisy Student ~\citep{sagawa2022uwilds} & $47.5 \pm 1.0$ & $32.1 \pm 0.8$  & - & - \\
            ICON & $50.6 \pm 1.3$ & $34.5 \pm 1.4$ & $90.1 \pm 0.4$ & $93.8 \pm 0.3$ \\
            Connect Later & $\mathbf{51.7 \pm 0.8}$ & $\mathbf{36.9 \pm 0.7}$ & $\mathbf{98.5 \pm 0.0}$ & $\mathbf{94.9 \pm 0.4}$\\
            \bottomrule
        \end{tabular}
    }
    \label{tbl:main-iwildcam}
\end{table*}
Tables~\ref{tbl:main-astro} and~\ref{tbl:main-iwildcam} compare the results of Connect Later with baseline methods. Connect Later outperforms all baselines on the OOD metric while maintaining comparable or better ID performance and achieves state-of-the-art performance on \iwildcam by 0.8\% OOD for ResNet-50, \camelyon by 1.1\% OOD for DenseNet121, and \classification by 3\%.

\paragraph{Connect Later improves OOD performance when \sft is minimally effective for UDA.} 
On \iwildcam, \sft minimally improves in OOD performance, while ERM+targeted augmentations improves by 6\% ID and OOD over both ERM and \sft. 
Connect Later improves over both \sft (by $6.8\%$) and ERM+targeted augs (by $0.9\%$) in OOD performance, indicating that the pretrained representations as well as the targeted augmentations are both important for OOD performance. Connect Later achieves a new state-of-the-art performance for ResNet-50 on the \iwildcam benchmark.

\paragraph{When \sft is effective, Connect Later still produces additional performance gains.}
On \camelyon, \classification, and \redshifts, Connect Later still outperforms all variants even though \sft already produces significant gains over ERM.
For \camelyon, \sft improves substantially over ERM in OOD average accuracy ($26.2\%$).
ERM+targeted augs outperforms \sft in ID accuracy by $4.4\%$, but does not improve OOD. 
Connect Later sets a new state-of-the-art on \camelyon with DenseNet121, improving on the best ID performance by 1.8\% (ERM+targeted augs) and OOD performance by 1.1\% (ICON). 

For \classification, \sft also performs significantly better than ERM: $7\%$ ID, $6.5\%$ OOD. 
ERM+targeted augs underperforms in ID accuracy compared to ERM ($-2.8\%$) and \sft ($-9.9\%$), likely due to the strong targeted augmentations. 
However, OOD accuracy of ERM+targeted augs is competitive with \sft, outperforming ERM. 
Connect Later outperforms the best baseline, \sft, by 12\% OOD and 2\% ID. The ID accuracy outperforms both \sft and ERM+targeted augs, showing a complementary benefit between pretraining and targeted augmentations. Connect Later sets a new state-of-the-art OOD performance on \classification by 3\% over Avocado, a heavily tuned random forest model with expert-designed features \citep{Boone_2019}.

\redshifts results are similar to \classification, with \sft significantly improving over ERM in both ID (7\% relative) and OOD (13\% relative) RMSE. 
Connect Later outperforms the best baseline variant, \sft, by 0.03 RMSE (11\% relative) with comparable ID error.

\paragraph{Connect Later improves OOD performance for CLIP fine-tuning.} We additionally evaluated the effectiveness of Connect Later for CLIP ViT-L/14 \citep{radford2021clip} on the \iwildcam dataset (Table~\ref{tbl:iwildcam-clip}). \Sft improves substantially over both ERM and ERM + targeted augs, likely due to CLIP's internet-scale pretraining dataset as well as the practical importance of pretraining for ViT performance. However, Connect Later still delivers additional gains over \sft (0.5\% ID, 1.4\% OOD).

\begin{table}[]
\caption{ID and OOD Macro F1 results for fine-tuning CLIP ViT-L on \iwildcam. Results are averaged over 15 trials and we report 95\% confidence intervals on each mean estimate. Rows with means within 1 interval of the best mean are bolded.}
    \centering
    
    \begin{tabular}{lcc}
    \toprule
    & ID Test & OOD Test \\
    \midrule
    ERM & $22.6 \pm 0.6$ & $7.8 \pm 0.2$ \\
    Standard fine-tuning & $\mathbf{55.3 \pm 1.4}$ & $42.8 \pm 0.9 $\\
    ERM + targeted augs & $23.8 \pm 0.7$ & $8.4 \pm 0.4$ \\
    Connect Later & $\mathbf{55.8 \pm 0.8 }$ & $\mathbf{44.2 \pm 0.9}$ \\
    \bottomrule
    \end{tabular}
    
    \label{tbl:iwildcam-clip}
\end{table}

\paragraph{Other baselines.}
DANN, CORAL, and Noisy Student did not produce competitive OOD average accuracy for either \iwildcam or \camelyon. 
ICON is the best baseline for \camelyon OOD average accuracy and is outperformed only by Connect Later.
For \classification and \redshifts, self-training improves both ID and OOD performance compared to ERM but underperforms \sft in both domains.


\subsection{Ablations}

We performed ablations on the model size, strength of pretraining augmentations (masking percentage for masked autoencoding), and LP-FT on \classification. 
We find that downstream performance is quite robust to masking percentage, while scaling up model size and LP-FT improve performance for pretrained models.

\begin{figure}
    \centering
    \includegraphics[scale=0.35]{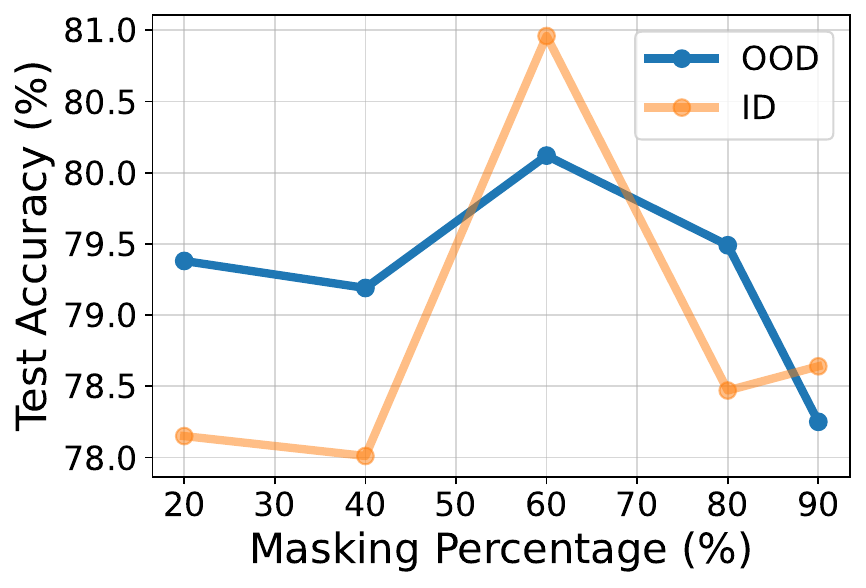}
    \caption{On the \classification task, Connect Later is relatively robust to pretraining masking percentage both ID and OOD, but 60\% masking performs best out of the percentages we tested.}
    \label{fig:ablations-masking}
\end{figure}


\paragraph{Model scale.} We tested Connect Later with a larger model ($\sim3\times$ the parameters of our model, $\text{21M} \rightarrow \text{69M}$), and find that the larger model produces higher ID and OOD accuracy (Table~\ref{tbl:ablations-size}). This suggests that scaling up the model is a promising way to further improve performance with Connect Later.

\paragraph{Strength of pretraining augmentations (masking percentage).} We vary the strength of pretraining augmentations with the MAE objective, as augmentation strength is parameterized solely by masking percentage. We tested pretraining masking percentages \ \{20, 40, 60, 80, 90\}\% with the same masking strategy (replace 10\% of masked indices with random values from the time-series, another 10\% are kept unchanged, and 80\% are replaced with the mask token, 0). We show the ID and OOD test accuracy of each variant in Figure~\ref{fig:ablations-masking}. Both ID and OOD performance peak at 60\% masking, although the performance of Connect Later is quite robust to the masking percentage. All masking percentages improve on OOD performance over \sft or ERM with targeted augmentations. Even the strongest augmentations (90\% masking) did not improve OOD performance over weaker augmentations. We hypothesize that strong generic augmentations may indiscriminately increase the connectivity between all source and target examples, including examples from different classes that should not be strongly connected.
\begin{table}[]
    \centering
    \caption{Scaling up model size of Connect Later produces improvements in both ID and OOD performance on the \classification task. }
    \begin{tabular}{@{}lcc@{}}
        \toprule
         Number of Parameters & ID Acc ($\uparrow$) & OOD  Acc ($\uparrow$)\\
         \midrule
        21M (default) & 78.47 & 79.49 \\
        69M & 80.38 & 80.55 \\
        \bottomrule
    \end{tabular}

    \label{tbl:ablations-size}
\end{table}

\begin{table}[]
    \centering
    \caption{Linear probing (LP) in addition to fine-tuning (FT) hurts performance for the ERM+targeted augs model but improves performance for Connect Later (tested on the \classification task).}
    \resizebox{0.45\textwidth}{!}{ 
        \begin{tabular}{@{}lcccc@{}}
            \toprule
             & \multicolumn{2}{c}{Connect Later} & \multicolumn{2}{c}{ERM+targeted augs} \\
             & ID Acc($\uparrow$) & OOD Acc($\uparrow$) & ID Acc($\uparrow$) & OOD Acc($\uparrow$) \\
             \midrule
            FT only & 78.07 & 78.6 & 77.88 & 68.43 \\
            LP-FT & 78.47 & 79.49 &  65.68	& 67.07  \\
            \bottomrule
        \end{tabular}
    }

    \label{tbl:ablations-LP}
\end{table}

\paragraph{Linear probing then fine-tuning.} \citet{kumar2022finetuning} showed that linear probing (with fixed neural embeddings) and then fine-tuning (LP-FT) the entire model improves both ID and OOD performance. Intuitively, full fine-tuning with a randomly initialized linear probe can destroy the pretrained features, and training the linear probe first mitigates this. 
We test LP-FT against FT only (all model weights are fine-tuned) with the Connect Later model and the ERM+targeted augs baseline.
We find that LP-FT improves OOD accuracy by 0.9\% over FT only when applied to Connect Later on \classification (Table~\ref{tbl:ablations-LP}).
On the other hand, LP-FT decreased OOD accuracy by 1.4\% when applied to ERM+targeted augs, which uses random initialization (no pretraining).
As a result, we use LP-FT on pretrained models but not on ERM or ERM+targeted augs.

\section{Discussion and Related Work}

\paragraph{Augmentations for pretraining.}
Data augmentations (e.g., cropping or masking) are vital to semi- and self-supervised learning objectives. Reconstructing a masked or noised input has been shown to produce useful pretrained representations across multiple modalities \citep{devlin2019bert, lewis2020bart, he2022mae, raffel2019exploring, chen2020simclr,he2020moco,caron2020swav}. 
In contrastive learning, models are trained to distinguish augmented ``views'' of the same input from views of a different input \citep{chen2020simclr, caron2020swav, he2020moco}. 
Our results demonstrate that though pretraining with generic augmentations alone produces inconsistent OOD performance across datasets, fine-tuning with targeted augmentations is able to better leverage these pretrained representations.

\paragraph{Augmentations for robustness.}
Data augmentation has been used to improve model robustness to label-independent changes (e.g. translation or rotation in vision) \citep{hendrycks2019augmix, rebuffi2021data, ng2020ssmba}. 
Existing augmentation strategies rely on generic perturbations that aim to increase the diversity of inputs \citep[e.g.,][]{simard2003best,krizhevsky2012imagenet,cubuk2019autoaugment,cubuk2020randaugment,devries2017improved,zhang2017mixup},
though prior work has shown that the type of data augmentations matters for performance \citep{chen2020simclr, xie2020unsupervised}.
Augmentations have also been leveraged in the self-training paradigm, which improves generalization to unseen data by training on the pseudo-labeled full dataset \citep{xie2020selftraining, sohn2020fixmatch, yang2021meanteacher}. 
We show that a self-training baseline with pseudo-labels from an ERM+targeted augs model does not outperform Connect Later, indicating that pretraining is important for robustness gains.
Connect Later exposes targeted augmentations as a design interface for improving robustness with knowledge of the distribution shift, while still leveraging pretrained representations.

\paragraph{Targeted augmentations.}
In domain shift problems, \citet{gao2023targeted} show that targeted augmentations designed with knowledge of the distribution shift outperform generic, or even target-aware (e.g. CutMix, \citet{yun2019cutmix}), augmentations on unseen data.
\citet{gao2023targeted} consider the domain generalization setting, in which the target dataset is unknown. We consider targeted augmentations in the domain adaptation setting, where we can model the target distribution with the unlabeled target data. 
In this setting, targeted augmentations provide the opportunity to naturally incorporate domain knowledge about the dataset and distribution shift.
In this work, we provide a general methodology for the design of such augmentations and show that targeted augmentations better leverage pretrained representations for complementary gains in OOD performance.
Certain aspects of the design process, such as the selection of feature space $z$ and transformation distribution $T$ could be learned from the unlabeled data itself, which we leave for future work. 


\section{Conclusion}

We show that pretraining with generic augmentations is not a panacea for all distribution shifts and tasks, and does not deliver consistent gains over supervised learning on labeled source data.
Pure supervised learning, however, does not use the unlabeled data or produce reusable representations. 
Connect Later allows for better leverage of pretrained representations for OOD performance by applying targeted augmentations at fine-tuning time. 

\section*{Impact Statement}
This paper presents work whose goal is to advance the field of Machine Learning. There are many potential societal consequences of our work, none of which we feel must be specifically highlighted here.

\section*{Acknowledgments}
We thank the anonymous reviewers for their comments, leading to substantial improvements of the paper. This research used resources of the National Energy Research Scientific Computing Center (NERSC), a Department of Energy Office of Science User Facility using NERSC award HEP-dessn.

\bibliography{iclr2024_conference}
\bibliographystyle{icml2024}

\newpage
\appendix
\onecolumn

\section{Additional Dataset Details}
\subsection{AstroClassification, Redshifts Datasets}
\label{app:data}

The \classification and \redshifts datasets were adapted from the 2019 Photometric LSST Astronomical Time-Series Classification Challenge \citep{theplasticcteam2018photometric} \footnote{https://zenodo.org/record/2539456}. This diverse dataset contains 14 types of astronomical time-varying objects, simulated using the expected instrument characteristics and survey strategy of the upcoming Legacy Survey of Space and Time \citep[LSST][]{ivezic2019lsst} conducted at the Vera C. Rubin Observatory. It includes two overall categories of time-series objects: \textit{transients}, short-lived events such as supernovae, and \textit{variable} sources, those with fluctuating brightness such as pulsating stars. Specifically, the dataset includes the following transients: type Ia supernovae (SNIa), SNIax, SNIa-91bg, SNIbc, SNII, superluminous supernovae (SLSN), tidal disruption events (TDE), and single lens microlensing events ($\mu$Lens-Single); and the following variable objects: active galactic nuclei (AGN), Mira variables, eclipsing binary systems (EB), and RR Lyrae (RRL).  

Millions of potential new objects are discovered per observing night, and important metadata such as object type, redshift, or other physical parameters, require astronomers to take time-intensive \textit{spectra} of each object. Spectra are a granular brightness vs. wavelength measurement at a single point in time, and are typically only taken for bright, nearby objects which require less exposure time than faint, faraway objects. The vast majority of discovered objects, however, will not have spectra but instead a time series of imaging data taken in 6 broad wavelength ranges, or \textit{photometric bands}. The time-varying behavior of these objects in these coarse wavelength bands does offer important clues about these physical parameters, but expert interpretation of spectra are traditionally required for confident labeling. Thus, our labeled training data for both \classification and \redshifts come from the unrepresentative subset of objects with spectra.

In these tasks, we are specifically interested in predicting the object type (e.g. type II supernova) and the cosmological redshift of objects in the unlabeled dataset. \textit{Cosmological redshift} is a proxy for distance in the universe, and an important piece of metadata for understanding an object's physical processes as well as other applications, such as estimating the expansion rate of the universe with type Ia supernovae. The redshift prediction task has been studied for individual object types, such as quasars \citep{nakoneczny2021photometric} and type Ia supernovae \citep{photozsnthesis}, but we consider a more realistic set of multiple object types.

\paragraph{Problem Setting.} 
The task is to predict object type for \classification (redshift for \redshifts) from time-series of object brightness. The input $x$ consists of flux measurements and associated uncertainties at times $\vt$ and photometric band that each measurement was taken in $\vb$: $\{F(t_i,b_j)\}_{i=1,j=1}^{T,W}, \{F_{\text{err}}(t_i, b_j)\}_{i=1,j=1}^{T,W}$. For this work, we map each $b \in \vb$ to the central wavelength of the $b$ band, which we denote $\vw$. The domain $d$ is binary, corresponding to whether the object has a spectrum (and thus a label). The labels $y$ are available only for objects with spectra, and are one of 14 types of astronomical time-varying objects for \classification (redshift of the object for \redshifts). We seek to optimize performance on the unlabeled data, which are generally fainter and further away than the labeled subset. We evaluate on these examples as well as held-out examples from the labeled subset.

\paragraph{Data.}
The training set of 7,846 objects is designed to emulate a sample of objects with spectra and thus biased toward brighter, more nearby objects compared to the test set of 3,492,888 objects. A random subset of 10,000 test set objects was selected for evaluation.
\begin{enumerate}
    \item \textbf{Source:} 6,274 objects
    \item \textbf{ID Test}: 782 objects
    \item \textbf{OOD Test:} 10,000 objects
\end{enumerate}
All data were simulated with the SuperNova ANAlysis \citep[SNANA,][]{kessler2009snana} software library. Further details about the astrophysical models and LSST instrument characteristics used in the simulation can be found in \citet{Kessler_2019}. 

\section{Data Augmentations}
\subsection{Generic Augmentations for Pretraining}

\paragraph{AstroClassification and Redshifts.} For the \classification and \redshifts datasets, we randomly mask a subset of the input sequence using the masked language modeling paradigm introduced by \cite{devlin2019bert}. Given an unlabeled input sequence $x$, a training input $x'$ can be generated by randomly masking elements of $x$ while the associated label $y$ consists of the original, unmasked values. The model is trained to use contextual information (unmasked elements) to successfully reconstruct most of the sequence. From our ablation experiments, we find that a masking percentage of 60\% produces the best downstream results. We follow an existing implementation for astronomical time-series \citep{astromer} and set 80\% of the masked elements to 0, replace 10\% with a random element from the sequence, and keep the remaining 10\% unchanged.

\paragraph{iWildCam and Camelyon17.} 
For \iwildcam, we use a ResNet-50 model pretrained on ImageNet with SwAV, a contrastive learning algorithm~\cite{caron2020swav}. 
For \camelyon, we use a DenseNet121 pretrained with SwAV on the unlabeled \camelyon dataset from \citet{sagawa2022uwilds}.
SwAV uses random cropping augmentations of different resolutions.

\subsection{Targeted Augmentations for Fine-Tuning}
\label{app:targeted-augs}

\paragraph{Redshifting for AstroClassification and Redshifts.} 

\begin{figure}
    \centering
    \includegraphics[scale=0.4]{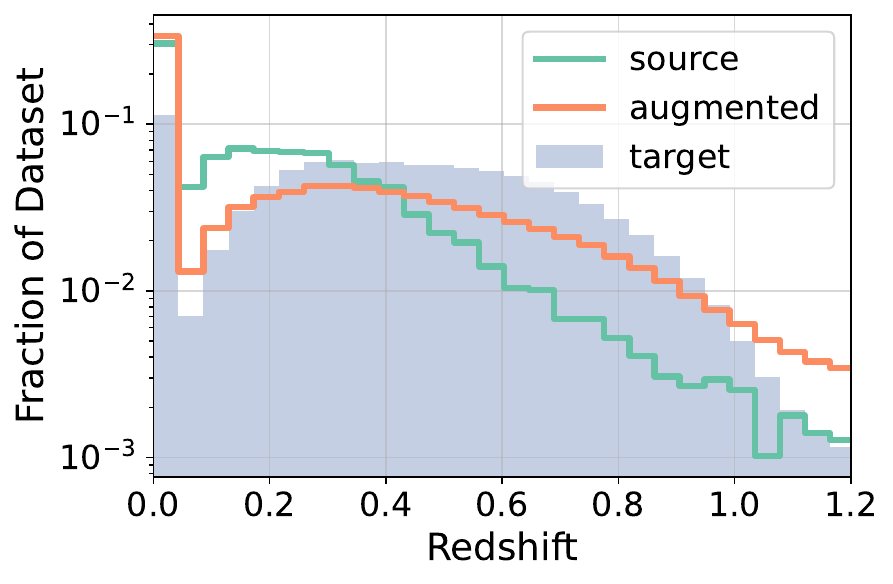}
    \caption{Redshift distributions of source, augmented, and target datasets for the \classification and \redshifts tasks.}
    \label{fig:z_dists}
\end{figure}

The OOD test set of the \classification and \redshifts datasets have many more high redshift objects than the source dataset, leading us to adopt an augmentation scheme to alleviate this shift. Figure~\ref{fig:z_dists} shows the redshift distributions of the source, augmented, and target datasets. Redshifting places each object at a new redshift and recomputes its light curve sampling, fluxes, and flux uncertainties accordingly. This augmentation algorithm was adapted from \citet{Boone_2019}. 

An input $\Xin \in \R^{T \times W}$ is a multivariate time series of flux values at specified times and observed wavelengths, $\{F(t_i, w_j)\}_{i=1,j=1}^{T,W}$.
We also have $\Xinerr \in \R^{T \times W}$, representing the flux errors corresponding to each element of $\mX$. We denote the elements of $\augXerr$ by $\{F_{\text{err}}(t_i, w_j)\}_{i=1,j=1}^{T,W}$.
Our goal is to model $F, F_{\text{err}}:\R\times\R \rightarrow \R$ at a new chosen redshift, $\znew$, to produce augmented inputs $\augX, \augXerr$.
\begin{itemize}
    \item We first construct a distribution from which to sample the new redshift, taking into account the current redshift of the object $\zorig$ as well as the target redshift distribution. We then sample a new redshift, $\znew \sim \text{loguniform}(0.95\zorig, \;\text{min}(1.5(1+\zorig)-1,\; 5\zorig))$.
    
    \item We fit a Gaussian process (GP) model for $F$ with training observations $\mX$ queried at the training input values $(\vt, \vw)$, and denote the predictive mean and variance of the GP as $F', F'_{\text{err}}$.

    \item Given the new redshift value $\znew$, we rescale the timestamps and wavelengths of the original observations to account for the physical effects of the new redshift value: $\vt_{\text{new}} = \frac{1+\znew}{1+\zorig}\vt$, $\vw_{\text{new}} = \frac{1+\znew}{1+\zorig}\vw$. We also randomly drop out 10\% as well as a large swath of $(\vt_{\text{new}}, \vw_{\text{new}})$ to simulate distinct observing seasons (telescope observing only occurs in the winter).

    \item We obtain GP predictions at test inputs $\{F'(t_{\text{new},i}, w_{\text{new},j})\}_{i=1,j=1}^{T,W}$, $\{F'_{\text{err}}(t_{\text{new},i}, w_{\text{new},i})\}_{i=1,j=1}^{T,W}$ and scale them by the log ratio of the new and original distances: 
    $$\tilde{\mX'}=10^{0.4(d(\znew)-d(\zorig))} \{F'(t_{\text{new},i}, w_{\text{new},j})\}_{i=1,j=1}^{T,W},$$ $$\tilde{\mX'}_{\text{err}}=10^{0.4(d(\znew)-d(\zorig))} \{F'_{\text{err}}(t_{\text{new},i}, w_{\text{new},j})\}_{i=1,j=1}^{T,W},$$ where $d(z)$ is the distance corresponding to redshift $z$.

    \item   We roughly model the observational noise of the telescope from the target data as a function of wavelength and sample $\epsilon \in \R^W$ from it. We define 
    $$\augX =\{\tilde{\augX}_{:,j} + \epsilon_j \}_{j=1}^{W}, \augXerr = \left\{\sqrt{\tilde{\mX'}_{\text{err},:,j}^2 + \epsilon_j^2} \right\}_{j=1}^{W}.$$

    \item We model the observational capabilities of the telescope to ensure that our augmented input $\augX, \augXerr$ does not fall below the threshold of detection. We ``accept" an augmented input $\augX, \augXerr$ if the signal-to-noise ratio (SNR) of at least two observations is over 5, i.e. $\text{SNR}(\augX_{i,j}, \mX'_{\text{err},i,j}) \geq 5$ for at least 2 of $i \in \{1,...,T\}, j \in \{1,...,W\}$. We define $\text{SNR}(x, x_{\text{err}})=\frac{|x|}{x_{\text{err}}}$.
\end{itemize}

\paragraph{Copy-Paste (Same Y) for iWildCam.}
This augmentation strategy randomizes the backgrounds of wildlife images to reduce the model's dependence on these spurious features for species classification. Specifically, a segmentation mask is applied to each image to separate the animal from the background, and the animal is ``copy-pasted" into a new background from a camera that has observed that animal species. This was the best performing augmentation strategy from \citet{gao2023targeted}.

\paragraph{Stain Color Jitter for Camelyon17.} 
This augmentation, originally from \citet{tellez2018whole}, alters the pixel values of the slide images to emulate different staining procedures used by different hospitals. The augmentation uses a pre-specified Optical Density (OD) matrix to project images from RGB space to
a three-channel hematoxylin, eosin, and DAB space before
applying a random linear combination. This was the best performing augmentation strategy from \citet{gao2023targeted}.

\section{Experimental Details}
\label{app:experiments}

\paragraph{AstroClassification and Redshifts.}
For \classification and \redshifts, we pretrain with a masked autoencoding objective:

\begin{align}
\label{eqn:mae_objective}
\sL_{\text{MAE}}(\encoder) = \E_{\inputx\sim \unlabeldist,\inputxp\sim\aug(\cdot \mid \inputx)}[(\encoder(\inputxp)-\inputx)^2]
\end{align}
We use an encoder-only Informer model \citep{zhou2021informer} with 8 encoder layers of 12 attention heads each. The model hidden dimension was chosen to be 768 and the layer MLPs have hidden dimension 256. Due to the 2-dimensional position data (each element of the time-series has an associated time and photometric band/wavelength) and irregular sampling of our dataset, we train a positional encoding based on learnable Fourier features following~\citet{li2021learnable}. We also select a random window of length 300 from each example (and zero-pad examples with fewer than 300 observations) to produce inputs of uniform shape. We perform pretraining with a batch size of 256 and learning rate 1e-4 (selected from 1e-3 $\sim$ 1e-6) for 75,000 steps. We finetune the pretrained model with linear probing for 20,000 steps (for pretrained models only) and learning rate 1e-4, then fine-tuning for 10,000 steps at learning rate of 4e-5. We increase the learning rate for models without pretraining to 1e-4 for FT. The \redshifts task uses LP learning rate of 5e-4 and FT learning rate of 1e-4. We decrease the learning rate per step with a linear scheduler.

\paragraph{iWildCam.} 
For pretraining, we use ResNet-50 pretrained on ImageNet with SwAV~\citep{caron2020swav}. During fine-tuning, we train all models for 15 epochs with early stopping on OOD validation performance, following~\citet{gao2023targeted}. For pretrained models, we also do 10 epochs of linear probing before fine-tuning \citep[LP-FT,][]{kumar2022finetuning} for 15 epochs, where the linear probe is trained with Adam and the linear probe weights used to initialize the fine-tuning stage is chosen with OOD validation performance. To reduce the noise in OOD results, for all methods we select the epoch in the last 5 epochs with the best OOD validation performance and report OOD test results with that version of the model. Following~\citet{gao2023targeted}, we allow for 10 hyperparameter tuning runs, where we sample the following hyperparameters independently from the following distributions: the linear probe learning rate ($10^{\text{Uniform}[-3,-2]}$), fine-tuning learning rate ($10^{\text{Uniform}[-5,-2]}$), and probability of applying the augmentation ($\text{Uniform}[0.5, 0.9]$) and pick the hyperparameter configuration with the best OOD validation performance. For ERM and ERM+targeted augmentations, we use the tuned hyperparameters from~\citet{gao2023targeted}. To decrease the confidence interval due to an outlier seed, the reported performance of Connect Later is averaged over 15 seeds. All other results are averaged over 5 seeds.

\paragraph{Camelyon17.} For pretraining, we use DenseNet121 pretrained on the unlabeled \camelyon dataset presented in \citet{sagawa2022uwilds} with SwAV \citep{caron2020swav}. During fine-tuning, we train all models for 15 epochs with early stopping on OOD validation performance, following~\citet{gao2023targeted}. For pretrained models, we also do 10 epochs of linear probing before fine-tuning \citep[LP-FT,][]{kumar2022finetuning} for 15 epochs, where the linear probe is trained with Adam and the linear probe weights used to initialize the fine-tuning stage is chosen with OOD validation performance. To reduce the noise in OOD results, for all methods we select the epoch with the best OOD validation performance and report OOD test results with that version of the model. Following~\citet{gao2023targeted}, we allow for 10 hyperparameter tuning runs, where we sample the following hyperparameters independently from the following distributions: the linear probe learning rate ($10^{\text{Uniform}[-3,-2]}$), fine-tuning learning rate ($10^{\text{Uniform}[-5,-2]}$), probability of applying the augmentation ($\text{Uniform}[0.5, 0.9]$), and augmentation strength ($\text{Uniform}[0.05, 0.1]$), and pick the hyperparameter configuration with the best OOD validation performance. All results are averaged over 20 seeds.

\section{Empirical Estimates of Connectivity}
\label{app:connectivity}

\begin{table}[]
    \centering
    \caption{Empirically estimated connectivity measures for \iwildcam, \classification, and \camelyon. \iwildcam and \classification results are averaged over 15 randomly selected class-domain pairs, while \camelyon results are averaged over all possible class-domain pairs.}
    \begin{tabular}{lccc}
        \toprule
         & across-domain & across-class & across-both \\ \midrule
        \iwildcam & 0.116 & 0.071 & 0.076 \\
        \classification & 0.287 & 0.159 & 0.097 \\
        \camelyon & 0.16 & 0.198 & 0.152 \\ 
        \bottomrule
    \end{tabular}
    \label{tbl:app-connectivity}
\end{table}

We empirically estimate connectivity measures for all of the datasets we tested on following the procedure outlined in Appendix D of \citet{shen2022connect}. 
Specifically, we train binary classifiers from scratch to predict the class-domain pair of a given input example. We randomly select 15 class-domain pairs for \iwildcam and \classification, while for \camelyon we use all class-domain pairs since \camelyon is a binary classification task. We label these class-domain examples following Appendix D of \citet{shen2022connect} and create a dataset with 80/10/10 train/validation/test split. We train using the same hyperparameters described in Appendix~\ref{app:experiments} for 3,000 steps with early stopping on the validation accuracy.
Our results are presented in Table~\ref{tbl:app-connectivity}.

\newcommand{\reachableset}{\sR}
\newcommand{\zpos}{z_{\posclass}}
\newcommand{\zneg}{z_{\negclass}}
\newcommand{\aaa}{a}
\newcommand{\bbb}{b}
\newcommand{\uone}{u_1}
\newcommand{\utwo}{u_2}

\section{Simple construction where Connect Later improves over pretraining or targeted augmentations alone}
\label{app:simple_example}

We give a simple construction for constrastive pretraining based on the construction in Proposition 3 (Appendix A.2) of~\citet{shen2022connect}, where Connect Later improves over pretraining (\sft) or targeted augmentations alone.

\paragraph{Data distribution.}
We consider binary classification with 2 domains.
Let $\sS = \{ \inputx \in \inputspace: \domainx = 1 \}$ and $\sT = \{ \inputx \in \sT: \domainx = 2\}$, and assume that $\sourcedist$ and $\targetdist$ are uniform over $\sS$ and $\sT$.
The unlabeled distribution for pretraining is the uniform distribution over $\inputspace$.
The source domain $\sS=\{1,2\}$ contains 2 points and the target domain $\sT=\{3,4,5,6,7,8\}$ contains 6 points.
For simplicity, we let the labels $\labelx$ be a deterministic function of the input $\inputx$.
The label space is $\labelspace = \{\negclass, \posclass\}$. The label for $\inputx\in \{1,3,5,7\}$ is $\labelx=\posclass$ and the label for $\inputx\in \{2,4,6,8\}$ is $\labelx=\negclass$.
Only the source data is labeled.

\paragraph{ERM with targeted augmentations.} ERM with targeted augmentations learns a model on source labeled data. To specialize to this section, the ERM objective is
\begin{align}
    \label{eq:erm-object-setup}
    \Lerm(\clf) = \E_{\inputx \sim \sourcedist, \inputx' \sim \augft(\cdot \mid \inputx)}[ \ell(\clf(\inputx'), y_x) ].
\end{align}
ERM returns a classifier $\empclferm \in \argmin_{\clf} \Lerm(\clf)$.

\paragraph{Spectral contrastive learning.}
Following \citet{haochen2021spectral} and \citet{shen2022connect}, we analyze contrastive learning from an augmentation graph perspective, where inputs $\inputx$ are connected via augmentations with edge weights $\pospairdist(\inputx, \inputxp)$, which represent the probability of $\inputx, \inputxp$ being a  positive pair (augmentations of the same input $\inputx$).
For theoretical analysis, we analyze the spectral contrastive learning objective:
\begin{align}
    \label{eq:scl}
        \lpretrain(\encoder) = -2 \cdot &\E_{(\inputx, \posx) \sim \pospairdist}\left[\encoder(\inputx)^\top \encoder(\posx)\right] +\E_{\inputx, \inputx' \sim \unlabeldist}\left[\left(\encoder(\inputx)^\top \encoder(\inputx')\right)^2\right].
\end{align}
The result of pretraining to optimize the above objective is an encoder $\empencoder: \inputspace \to \R^\embeddim$.

\begin{figure}[t]
    \centering
	\includegraphics[width=0.6\linewidth]{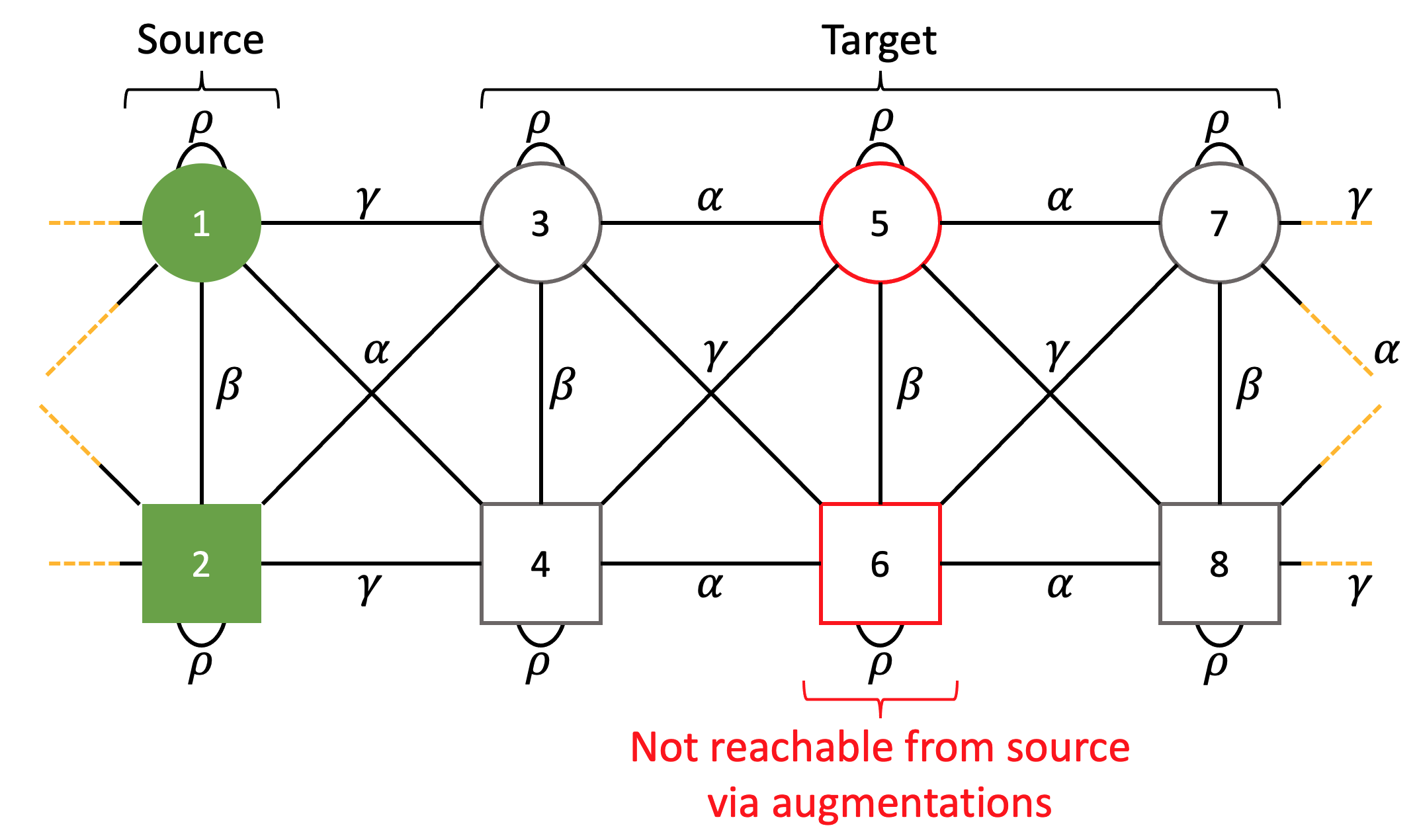}
	\caption{%
    Example distribution of data and augmentations for contrastive learning where Connect Later improves OOD performance over contrastive pretraining+standard fine-tuning and ERM+targeted augmentations. The augmentation graph is similar to~\citet{shen2022connect} except the edge weights connecting 1,2 and 3,4 are swapped. The shapes represent classes, while the labeled data is shaded in green. The generic augmentation probabilities are marked as edge weights, where we assume that $\alpha > \gamma + \beta$. Here, targeted augmentations which first swap inputs 1 and 2 before applying a generic augmentation help to align the source and target. However, some target inputs are not reachable via augmentations from source inputs. \Sft can generalize throughout the target domain, but only in conjunction with targeted augmentations that align the source and target. The orange dotted lines on the far ends connect to each other (the graph wraps around).}
    \label{fig:simple_example}
\end{figure}

\paragraph{Linear probing (fine-tuning step).}
Instead of analyzing fine-tuning, we follow~\citet{shen2022connect} and analyze linear probing on top of the pretrained representations from the encoder.
We train a linear model with parameters $\linmat \in \R^{\numcls \times \embeddim}$, where $\numcls$ is the number of classes.
We minimize the objective:
\begin{align}
    \label{eq:sq-loss}
    \lfinetune(\linmat) = \E_{\inputx \sim \sourcedist}\left[ \ell(\linmat \empencoder(\inputx), \labelx) \right] + \eta \| \linmat \|_F^2,
\end{align}
where $\ell$ is the squared loss and we take $\labelx\in \R^\embeddim$ to be a one-hot encoding of the class label.
The resulting classifier is $\empclf(\inputx) = \argmax_{i \in [\numcls]} (\emplinmat \empencoder(\inputx))_i$.

\paragraph{Pretraining augmentations (Figure~\ref{fig:simple_example})}
We define the pretraining augmentation distribution $\aug(\cdot \mid \inputx)$ to be
\begin{align}
        \aug(\inputxp \mid \inputx) =
    \begin{cases}
        \augprobr & \inputx = \inputxp\\
        \augproba & \{x',x\}\in \{\{1,4\}, \{3,5\}, \{5,7\}, \{2,5\}, \{4,6\}, \{6,8\}, \{1, 8\}, \{2, 7\}\\
        \augprobb & \{\inputxp, \inputx\} \in \{ \{1, 2\}, \{3, 4\}, \{5, 6\}, \{7, 8\} \} \\ 
        \augprobg & \{\inputxp, \inputx\} \in \{ \{1, 3\}, \{2, 4\}, \{3, 6\}, \{4, 5\}, \{5, 8\}, \{6, 7\}, \{1,7\}, \{2,8\} \} \\  
    \end{cases}. 
\end{align}
Notice that the weight between 1,3 is $\augprobg$ and the weight between 1,4 is $\augproba$, and the weights are similarly swapped for 2,4, and 2,5. 
We assume that $\augprobr, \augproba, \augprobb$, and $\augprobg$ are in $(0, 1)$ and are distinct. We also assume that the augmentation probabilities satisfy $\augprobr > \max\{\augproba, \augprobb\}$ and $\min\{\augproba, \augprobb\} > \augprobg$.
Following~\citet{shen2022connect}, we can convert these to positive pair probabilities $\pairprobr,\pairproba,\pairprobb,\pairprobg$ with similar properties by renormalizing.

Given the above setting, the following is a simplified form of Proposition 3 from~\citet{shen2022connect}, if we instead use the following augmentation distribution, which swaps the edge weight magnitudes that involve nodes 1 and 2:
\begin{align}
        \sA_{\text{prop}}(\inputxp \mid \inputx) =
    \begin{cases}
        \augprobr & \inputx = \inputxp\\
        \augproba & \{x',x\}\in \{\{1,3\}, \{3,5\}, \{5,7\}, \{2,4\}, \{4,6\}, \{6,8\}, \{1, 7\}, \{2, 8\}\\
        \augprobb & \{\inputxp, \inputx\} \in \{ \{1, 2\}, \{3, 4\}, \{5, 6\}, \{7, 8\} \} \\ 
        \augprobg & \{\inputxp, \inputx\} \in \{ \{1, 4\}, \{2, 3\}, \{3, 6\}, \{4, 5\}, \{5, 8\}, \{6, 7\}, \{1,8\}, \{2,7\} \} \\  
    \end{cases}. 
\end{align}
\begin{proposition}[\citet{shen2022connect}]
    \label{prop:separation}
     With the above construction for the input space $\inputspace$, unlabeled distribution $\unlabeldist$, and data augmentation $\sA_{\text{prop}}$, for some feature dimension $k \in \Z^+$ a linear probe trained on contrastive pre-trained features achieves 0 target error:
     $\Lzeroone(\empclf) = 0$.
    However, for all $\embeddim\in\Z^+$, there exists a minimizer $\empclf_\text{erm}$ of the ERM objective (with data augmentations according to $\sA_{\text{prop}}$) that has non-zero error: $\Lzeroone(\empclf_\text{erm}) = 1/3$.
\end{proposition}  

\paragraph{ERM with targeted augmentations can get high OOD error.}
In general, we proceed by defining the following targeted augmentation, which allows us to reduce to the setting of Proposition~\ref{prop:separation}:
\begin{align}
    \label{eqn:targeted_aug}
        \augft(\inputxp \mid \inputx) =
    \begin{cases}
        1 & \{\inputxp, \inputx\} \in \{1,4\}, \{2,3\} \\
        1 & \inputx=\inputxp \text{ and } \inputx \notin \{1,2\}\\
        0 & \text{otherwise}
    \end{cases}
\end{align}
which transforms input 1 to 4 and the input 2 to 3, while keeping all other inputs the same.
Since the ERM with augmentations objective will not contain a term involving inputs 5,6,7, or 8 and thus the prediction on these inputs do not affect the objective, there exists a minimizer of the ERM objective (Equation~\ref{eq:erm-object-setup}) that predicts the wrong label for inputs 5,6,7,8 and has target error 2/3.
This is because these nodes are unreachable via augmentations of the source inputs, and thus the ERM objective can be minimized with any arbitrary prediction on these inputs.

\paragraph{\Sft has high OOD error.}
By Proposition~\ref{prop:separation}, \sft after contrastive pretraining has zero target (OOD) error when the pretraining augmentations do not have swapped edges.
By symmetry, \sft (contrastive pretraining + linear probing) on our augmentation graph with pretraining augmentations $\aug$ outputs the opposite label for all target inputs, resulting in an OOD error of 1. This is because the source and target domains are misaligned in our augmentation graph.

\paragraph{Connect Later achieves zero OOD error.}
Connect Later applies targeted augmentations $\augft$ during the linear probing step (on top of contrastive pretrained representations). This choice of targeted augmentations reduces to the setting of Proposition~\ref{prop:separation} where the labeled source domain consists of the inputs 3,4 instead. By the symmetry of the graph and applying Proposition~\ref{prop:separation}, Connect Later achieves 0 OOD error.


\end{document}

%% file: macros.tex
\usepackage{bm}

\def\vb{{\bm{b}}}

\def\vt{{\bm{t}}}

\def\vw{{\bm{w}}}

\def\mX{{\bm{X}}}

\newcommand{\loss}{\ell}
\newcommand{\losstext}{\text{loss}}
\newcommand{\pstar}{p^*}

\newcommand{\classification}{\textsc{AstroClassification}\xspace}
\newcommand{\redshifts}{\textsc{Redshifts}\xspace}
\newcommand{\iwildcam}{\textsc{iWildCam-WILDS}\xspace}
\newcommand{\camelyon}{\textsc{Camelyon17-WILDS}\xspace}
\newcommand{\sft}{standard fine-tuning\xspace}
\newcommand{\Sft}{Standard fine-tuning\xspace}

\newcommand{\augX}{\mX'}
\newcommand{\augXerr}{\mX'_{\text{err}}}
\newcommand{\Xin}{\mX}
\newcommand{\Xinerr}{\mX_{\text{err}}}
\newcommand{\znew}{z'}
\newcommand{\zorig}{z_{\text{orig}}}

\newcommand{\sourcedist}{P_S}
\newcommand{\targetdist}{P_T}
\newcommand{\unlabeldist}{P_U}

\newcommand{\inputx}{x}

\newcommand{\labelx}{y_x}
\newcommand{\inputxp}{{x'}}

\newcommand{\domainx}{d_x}

\newcommand{\numcls}{r}

\newcommand{\embeddim}{k}

\newcommand{\posx}{\inputx^+}
\newcommand{\pospairdist}{S_+}
\newcommand{\dattract}{d_+}
\newcommand{\drepel}{d_-}

\newcommand{\inputspace}{\sX}
\newcommand{\labelspace}{\sY}

\newcommand{\aug}{\mathcal{A}_{\text{pre}}}
\newcommand{\augft}{\mathcal{A}_{\text{ft}}}

\newcommand{\augproba}{{\alpha'}}
\newcommand{\augprobb}{{\beta'}}
\newcommand{\augprobg}{{\gamma'}}
\newcommand{\augprobr}{{\rho'}}

\newcommand{\pairproba}{{\alpha}}
\newcommand{\pairprobb}{{\beta}}
\newcommand{\pairprobg}{{\gamma}}
\newcommand{\pairprobr}{{\rho}}

\newcommand{\posclass}{1}
\newcommand{\negclass}{-1}

\newcommand{\lossft}{\losstext_{\text{ft}}}

\newcommand{\encoder}{\phi}
\newcommand{\empencoder}{\widehat{\encoder}}

\newcommand{\head}{h}

\newcommand{\emphead}{\widehat{\head}}

\newcommand{\clf}{f}
\newcommand{\empclf}{\widehat{\clf}}
\newcommand{\linmat}{B}
\newcommand{\emplinmat}{\widehat{\linmat}}
\newcommand{\empclferm}{\widehat{\clf}_\text{erm}}

\newcommand{\lpretrain}{\sL_\text{pretrain}}

\newcommand{\lfinetune}{\sL}

\newcommand{\Lzeroone}{\sL_{0-1}}
\newcommand{\Lerm}{\sL_\text{ERM}}

\DeclareMathOperator*{\argmin}{arg\,min}
\DeclareMathOperator*{\argmax}{arg\,max}

\newlength{\widebarargwidth}
\newlength{\widebarargheight}
\newlength{\widebarargdepth}

%% file: std_macros.tex
\newcommand\sA{\ensuremath{\mathcal{A}}}

\newcommand\sL{\ensuremath{\mathcal{L}}}

\newcommand\sR{\ensuremath{\mathcal{R}}}
\newcommand\sS{\ensuremath{\mathcal{S}}}
\newcommand\sT{\ensuremath{\mathcal{T}}}

\newcommand\sX{\ensuremath{\mathcal{X}}}
\newcommand\sY{\ensuremath{\mathcal{Y}}}
\newcommand\sZ{\ensuremath{\mathcal{Z}}}

\newcommand\R{\ensuremath{\mathbb{R}}} 
\newcommand\Z{\ensuremath{\mathbb{Z}}} 

\ifthenelse{\isundefined{\definition}}{\newtheorem{definition}{Definition}}{}
\ifthenelse{\isundefined{\assumption}}{\newtheorem{assumption}{Assumption}}{}
\ifthenelse{\isundefined{\hypothesis}}{\newtheorem{hypothesis}{Hypothesis}}{}
\ifthenelse{\isundefined{\proposition}}{\newtheorem{proposition}{Proposition}}{}
\ifthenelse{\isundefined{\theorem}}{\newtheorem{theorem}{Theorem}}{}
\ifthenelse{\isundefined{\lemma}}{\newtheorem{lemma}{Lemma}}{}
\ifthenelse{\isundefined{\corollary}}{\newtheorem{corollary}{Corollary}}{}
\ifthenelse{\isundefined{\alg}}{\newtheorem{alg}{Algorithm}}{}
\ifthenelse{\isundefined{\example}}{\newtheorem{example}{Example}}{}
\newcommand{\E}{\ensuremath{\mathbb{E}}} 